\documentclass[runningheads]{llncs}
\usepackage[T1]{fontenc}
\usepackage{graphicx}
\usepackage{amsmath}
\usepackage{amsfonts}
\usepackage{hyperref}
\usepackage{bm}
\usepackage{booktabs}
\usepackage{multirow}
\usepackage[misc]{ifsym}
\newcommand{\corr}{(\Letter)}
\usepackage{mwe}

\begin{document}

\title{Bridging Logic and Learning: Decoding Temporal Logic Embeddings via Transformers}

\titlerunning{Decoding Temporal Logic Embeddings via Transformers}

\author{Sara Candussio\inst{1}\orcidID{0009-0004-5198-6970} \corr \and
Gaia Saveri\inst{1}\orcidID{0009-0003-2948-7705} \and
Gabriele Sarti\inst{2}\orcidID{0000-0001-8715-2987} \and Luca Bortolussi\inst{1}\orcidID{0000-0001-8874-4001}}

\authorrunning{S. Candussio et al.}

\institute{AILab, MIGe, University of Trieste, 34127 Trieste, IT \email{\{sara.candussio,gaia.saveri\}@phd.units.it}, \email{lbortolussi@units.it}
\and
Center for Language and Cognition (CLCG), University of Groningen \email{g.sarti@rug.nl}}

\tocauthor{Sara Candussio, Gaia Saveri, Gabriele Sarti, Luca Bortolussi} 
\toctitle{Bridging Logic and Learning: Decoding Temporal Logic Embeddings via Transformers}

\maketitle              

\begin{abstract}
Continuous representations of logic formulae allow us to integrate symbolic knowledge into data-driven learning algorithms. If such embeddings are semantically consistent, i.e. if similar specifications are mapped into nearby vectors, they enable continuous learning and optimization directly in the semantic space of formulae. However, to translate the optimal continuous representation into a concrete requirement, such embeddings must be invertible. We tackle this issue by training a Transformer-based decoder-only model to invert semantic embeddings of Signal Temporal Logic (STL) formulae. 
STL is a powerful formalism that allows us to describe properties of signals varying over time in an expressive yet concise way. By constructing a small vocabulary from STL syntax, we demonstrate that our proposed model is able to generate valid formulae after only $1$ epoch and to generalize to the semantics of the logic in about $10$ epochs. Additionally, the model is able to decode a given embedding into formulae that are often simpler in terms of length and nesting while remaining semantically close (or equivalent) to gold references.
We show the effectiveness of our methodology across various levels of training formulae complexity to assess the impact of training data on the model's ability to effectively capture the semantic information contained in the embeddings and generalize out-of-distribution.
Finally, we deploy our model for solving a requirement mining task, i.e. inferring STL specifications that solve a classification task on trajectories, performing the optimization directly in the semantic space.  

\end{abstract}

\section{Introduction}

Integrating learning algorithms and symbolic reasoning is an increasingly prominent research direction in modern Artificial Intelligence (AI), towards striking a balance between the high efficiency of black box data-driven Machine Learning (ML) models and the interpretability of logical languages. The cornerstone of such efforts, within the Neuro-Symbolic (NeSy) computing paradigm~\cite{nesy-wave,marra2024}, is knowledge representation by means of formal languages rooted on logic. A promising approach to combine logic with machine learning is that of mapping logic formulae into continuous vectors, i.e. embeddings, which can be natively integrated into ML algorithms~\cite{pooling-logic,robot-stl,contrastive-logic,stl2vec,embedding-gnn}. 

Having real-valued representation of logic specifications preserving their semantic, i.e. mapping similar formulae to nearby vectors, enables continuous learning and optimization directly in the semantic space of formulae, e.g. for conditioning generative models on producing data compliant to some background knowledge~\cite{stl2vec} or finding a requirement able to discriminate among regular and anomalous points~\cite{semantic-db}. However, to translate the devised optimal continuous representation into a concrete requirement, hence promoting interpretability and reliability of the resulting system, such embeddings must be invertible. 

In this work we focus on temporal data and on a dialect of Linear Temporal Logic, namely Signal Temporal Logic (STL)~\cite{stl,temporal-logic}, which is emerging as the de-facto standard language for stating specifications of continuous systems varying over time, in various domains such as biological or cyber-physical systems~\cite{stl-cps}. Indeed, STL is powerful enough to describe many phenomena, yet easily interpretable, as it avoids the vagueness and redundancy of natural language, still being easy to translate in common words~\cite{deep-stl}. For example, in STL one can state properties like "within the next $10$ minutes, the temperature will reach at least $25$ degrees, and will stay above $22$ degrees for the next hour". Notably, there exists a well-defined procedure for consistently embedding STL specifications in a real vector space, grounded directly in the semantics of the logic via kernel-based methods~\cite{stl-kernel,stl2vec}. This embedding, however, is not invertible, essentially due to the fact that semantically equivalent formulae with different syntactic structure are mapped in the same vector.  

On the other hand, Language Models (LMs) based on the Transformer architecture~\cite{transformer} have reached astonishingly high performance on a wide range of applications and for different data modalities~\cite{lm-survey}. Thanks to their effective and efficient self-attention mechanism, LMs have proven to be the most-powerful general-purpose representation learning models, setting new standards on various domain, beyond the traditional Natural Language Processing (NLP). For this reason, we believe that a decoder-only Transformer-based model could prove itself an effective choice for inverting continuous representations of STL formulae. Indeed, we tackle such decoding task as a translation from semantic vectors to strings representing formulae, constructing a small vocabulary from STL syntax. 

\paragraph{Our contributions} consist in: (i) end-to-end training of a decoder-only Transformer model for the downstream task of reconstructing STL specifications from a continuous representation encoding their semantics (Section~\ref{subsec:inversion}); (ii) extensive experimental analysis on the effectiveness of our methodology across various levels of training formulae complexity to assess the impact of training data on the model's ability to effectively capture the semantic information contained in the embeddings and generalize out-of-distribution, as well as comparisons to related approaches (Section~\ref{subsec:inversion}); (iii) leveraging the proposed architecture for requirement mining, i.e. inferring formally specified system properties from observed behaviors, by integrating our architecture inside a Bayesian Optimization (BO) loop, proving that the resulting model is able to extract interpretable information form the input data, promoting knowledge discovery about the system (Section~\ref{subsec:reqmining}). Data, code and trained models presented in this work can be found at can be found at \href{https://github.com/gaoithee/transformers/tree/main/src/transformers/models/stldec}{this}\footnote{\url{https://github.com/gaoithee/transformers/tree/main/src/transformers/models/stldec}} link.




\section{Related Work}\label{sec:relatedworks}
\paragraph{The ability of LM to understand formal languages} in general, and temporal logic in particular, has been explored in the literature under different perspectives. A number of works exploits Transformed-based architectures to translate informal natural language statements to formal specifications, towards aiding the error-prone and time-consuming process of writing requirements~\cite{nl2formal-survey}. In this context, either an encoder-decoder LM is trained from scratch as accurate translator from unstructured natural language sentences to STL formulae~\cite{deep-stl}, or off-the-shelf LMs are finetuned on an analogous translation task for first-order logic (FOL) or linear-time temporal logic (LTL)~\cite{nl2spec,nlp2formal}. Notably, Transformers are also trained to solve tasks typically pertaining to the formal methods realm: in~\cite{transformer-ltl} a LM is trained end-to-end for solving the LTL verification problem of generating a trace satisfying a given formula, showing generalization abilities of the model to the semantics of the logic; taking a dual perspective, in~\cite{mamba-ltl} the LTL requirement mining problem is framed as an auto-regressive language modeling task, in which an encoder-decoder LM is trained as a translator having as source language the input trace and as target language LTL. 

Finally, performing symbolic regression (SR) using Transformer-based models is gaining momentum in the field as an alternative to genetic programming. SR is indeed the problem of inferring a symbolic mathematical expression of a function of which we have collected some observations in the form of input-output pairs: in \cite{transformer-sr} a LM is trained to simultaneously predict the skeleton and the numerical constants of the searched expression, possibly augmenting the generation with a planning strategy~\cite{transformer-sr-planning}; in~\cite{odeformer} the investigation is pushed even further, as a Transformer is trained to infer multidimensional ordinary differential equations from observations, i.e. to model a dynamical systems from data.

\paragraph{Mining STL specifications from data} has seen a surge of interest in the last few years, as reported in~\cite{stl-mining-survey}. Many of such works decompose the requirement mining task in two steps, i.e. as bi-level optimization problem: they first learn the structure of the specification from data, and then instantiate a concrete formula using parameter inference methods~\cite{gp-two,decision-tree,enumerate-stl,roge,gp-one,grid-based}. Both in~\cite{roge,semantic-db} Bayesian optimization, and in particular GP-UCB, is used towards optimizing the searched specification. In the same spirit of this work, \cite{semantic-db} learns simultaneously both the structure and the parameters of the STL requirement, by performing optimization in a continuous space representing the semantics of formulae. 





\section{Background}\label{sec:background}

\paragraph{Signal Temporal Logic} (STL) is a linear-time temporal logic which expresses properties on trajectories over dense time intervals \cite{stl}. Signals (or trajectories) are here defined as functions $\xi: I\rightarrow D$, where $I\subseteq \mathbb{R}_{\geq 0}$ is the time domain and $D\subseteq \mathbb{R}^k, k\in \mathbb{N}$ is the state space. The syntax of STL is given by:
$$\varphi:=tt\mid\pi\mid\lnot\varphi\mid \varphi_1\land\varphi_2\mid\varphi_1\mathbf{U}_{[a, b]}\varphi_2$$
where $tt$ is the Boolean \emph{true} constant; $\pi$ is an \emph{atomic predicate}, i.e.\ a function over variables $\bm{x}\in \mathbb{R}^n$ of the form $f_{\pi}(\bm{x})\geq 0$;
$\lnot$ and $\land$ are the Boolean \emph{negation} and \emph{conjunction}, respectively (from which the \emph{disjunction} $\lor$ follows by De Morgan's law); $\mathbf{U}_{[a, b]}$, with $a, b \in \mathbb{Q}, a<b$, is the \emph{until} operator, from which the \emph{eventually} $\mathbf{F}_{[a, b]}$ and the \emph{always} $\mathbf{G}_{[a, b]}$ temporal operators can be derived. We can intuitively interpret the temporal operators over $[a, b]$ as follows: a property is \textit{eventually} satisfied if it is satisfied at some point inside the temporal interval, while a property is \textit{globally} satisfied if it is true continuously in $[a, b]$; finally the \textit{until} operator captures the relationship between two conditions $\varphi, \psi$ in which the first condition $\varphi$ holds until, at some point in $[a, b]$, the second condition $\psi$ becomes true. We call $\mathcal{P}$ the set of well-formed STL formulae. STL is endowed with both a \emph{qualitative} (or Boolean) semantics, giving the classical notion of satisfaction of a property over a trajectory, i.e. $s(\varphi, \xi, t) = 1$ if the trajectory $\xi$ at time $t$ satisfies the STL formula $\varphi$, and a  \emph{quantitative} semantics, denoted by $\rho(\varphi, \xi, t)\in \mathbb{R}$. The latter, also called \emph{robustness}, is a measure of how robust is the satisfaction of $\varphi$ w.r.t. perturbations of the signals. Intuitively, robustness measures how far is a signal $\xi$ from violating a specification $\varphi$, with the sign indicating the satisfaction status. Indeed, robustness is compatible with satisfaction via the following \emph{soundness} property: if $\rho(\varphi, \xi, t) > 0$ then $s(\varphi, \xi, t) = 1$ and if $\rho(\varphi, \xi, t) < 0$ then $s(\varphi, \xi, t) = 0$. When $\rho(\varphi, \xi, t) = 0$ arbitrary small perturbations of the signal might lead to changes in satisfaction value. 

\paragraph{Embeddings of STL formulae} are devised with an ad-hoc kernel in \cite{stl-kernel}, which yields representations that have been experimentally proven to be semantic-preserving~\cite{stl2vec}, i.e. STL specifications with similar meaning are mapped to nearby vectors. Indeed, such embeddings are not learned but grounded in the semantics of the logic, moving form the observation that robustness allows formulae to be considered as functionals mapping trajectories into real numbers, i.e.\ $\rho(\varphi,\cdot): \mathcal{T}\rightarrow \mathbb{R}$ such that $\xi\mapsto \rho(\varphi, \xi)$. Considering these as feature maps, and fixing a probability measure $\mu_0$ on the space of trajectories $\mathcal{T}$, a kernel function capturing similarity among STL formulae on mentioned feature representations can be defined as:
\begin{equation}
k(\varphi, \psi) = \langle \rho(\varphi, \cdot), \rho(\psi, \cdot) \rangle = \int_{\xi\in \mathcal{T}} \rho(\varphi, \xi) \rho(\psi, \xi) d\mu_0(\xi)
\label{eq:stl-kernel}
\end{equation}
opening the doors to the use of the scalar product in the Hilbert space $L^2$ as a kernel for $\mathcal{P}$; at a high level, this results in a kernel having high positive value for formulae that behave similarly on high-probability trajectories (w.r.t. $\mu_0$), and viceversa low negative value for formulae that on those trajectories disagree. Hence, a $D$-dimensional embedding $k(\varphi)\in \mathbb{R}^D$ of a formula $\varphi$ is obtained from Equation~\ref{eq:stl-kernel} by fixing (possibly at random) an \emph{anchor set} of $D$ STL requirements $\psi_1, \ldots, \psi_D$ such that $k(\varphi) = [k(\varphi, \psi_1), \ldots, k(\varphi, \psi_D)]$. 

\paragraph{Transformers} are a class of deep learning models designed to autoregressively handle sequential data effectively by leveraging self-attention mechanism~\cite{transformer}. Given a sequence of tokens $t=[t_1, \ldots, t_m]$, the model learns the distribution $P(t)$ that can be decomposed as $P(t) = P(t_1)\prod_{i=1}^{m-1}P(t_{i+1}|t_1, \ldots, t_i)$ using the chain rule, so that each conditional can be parameterized using a neural network optimized via Stochastic Gradient Descent (SGD) to maximize the likelihood of a corpus used for training.

Here, we focus on the decoder-only variant of this architecture~\cite{gpt}, whose objective is to predict the next token given a context made of the last $k$ already generated tokens. The fundamental learning mechanism in the Transformer decoder is the attention mechanism, that allows the model to contextualize token representation at each layer, dynamically determining the importance of each token in a computationally efficient (i.e. parallelizable) way.

\section{Decoding STL Embeddings via Transformers}\label{sec:method}

The main research question addressed with this work is that of checking whether a Transformer-based decoder-only model is able to decode a STL formula starting from a real-valued vector representing its semantics. In this context, we frame this problem as the approximate inversion of STL kernel embeddings computed as described in Section~\ref{sec:background}, i.e. using Equation~\ref{eq:stl-kernel}. A positive answer to this question can be interpreted as an experimental proof of the fact that, in this setting, ad-hoc trained Transformers are able to grasp the semantics of STL. 

\subsection{Data}\label{subsec:data}

In order to train the model, we need to collect a set of pairs $\{(\varphi_i, k(\varphi_i))\}_i$ consisting of STL formulae $\varphi_i$ and their corresponding kernel embeddings $k(\varphi_i)$ computed as described in Section~\ref{eq:stl-kernel}. To construct such a set of formulae $\varphi$, we can consider a distribution $\mathcal{F}$ over STL formulae, defined by a syntax-tree random recursive growing scheme, which recursively generates the nodes of a formula given the probability $p_{\text{leaf}}$ of each node being an atomic predicate, and a uniform distribution over the other operator nodes. Intuitively, the higher $p_{\text{leaf}}$, the smaller (in terms of number of operators and depth of the resulting syntax tree) the generated formula will be. 

\begin{figure}
    \centering
    \includegraphics[width=0.75\linewidth]{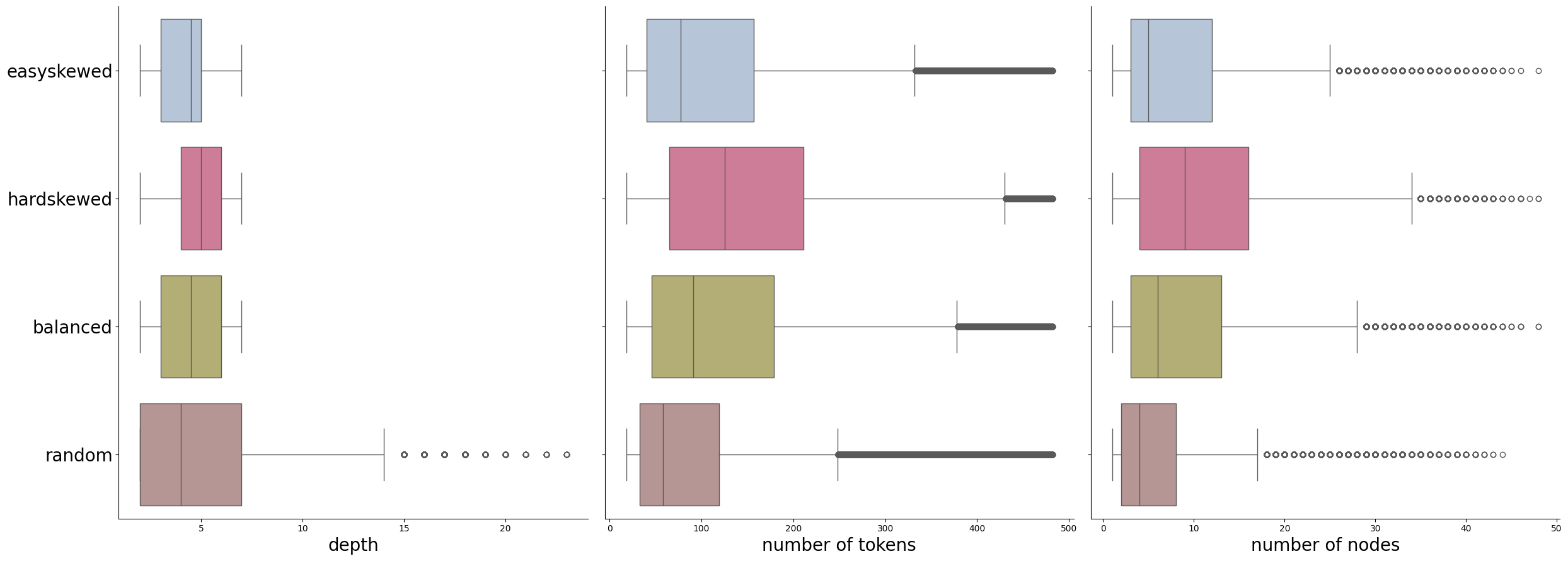}
    \caption{Distribution of depth, number of tokens and number of nodes of the formulae composing the different training sets used.}
    \label{fig:structure-training}
\end{figure}

The work of \cite{qin2024sometimes} spots a light on the impact that training data has in making the model learn hierarchical patterns, resulting in different generalization abilities at inference time. While they work on linear and hierarchical rules on natural language syntax trees, our case study is focused on syntactically reasonable representations of semantic vectors. We question whether, also in this different scenario, complexity level of training data (i.e. the composition of these sets in terms of formulae depths) should be reflected in some margin into the model's prediction abilities.

Specifically, we compose $4$ different training sets, by sampling from the distribution $\mathcal{F}$ varying the parameter $p_{\text{leaf}}$ and possibly filtering the obtained formulae based on the desired depth, each containing $78000$ examples as follows:

\begin{itemize} 
\item \texttt{easyskewed}: $52000$ easier formulae (i.e. of depths $2$, $3$, $4$), $26000$ harder formulae (i.e. of depths $5$, $6$, $7$); 
\item \texttt{hardskewed}: $26000$ easier formulae (i.e. of depths $2$, $3$, $4$), $52000$ harder formulae (i.e. of depths $5$, $6$, $7$); 
\item \texttt{balanced}: $13000$ formulae of each considered depth level (i.e. depths $2$, $3$, $4$, $5$, $6$, $7$); 
\item \texttt{random}: $78000$ formulae randomly sampled from $\mathcal{F}$ with $p_{\text{leaf}} = 0.45$, without filtering on the minimum or maximum depths; the result is a broadly spread distribution, comprising formulae from depth $2$ to depth $23$. 
\end{itemize}

A glance on the distribution of the depth and number of nodes of the syntax tree of the formulae in the different training sets is shown in Figure~\ref{fig:structure-training}.

The idea behind the exploitation of multiple training sets is to test whether the model can effectively learn to reconstruct a formula corresponding to a given embedding, or if the learning process is limited to replicating a superficial pattern observed in the training data~\cite{qin2024sometimes}. To assess the generalization abilities acquired during the training stage, we test each model on a balanced test set consisting in formulae with various depths (from $2$ to $7$).

\subsection{Model}\label{subsec:architecture}
The model follows a decoder-only architecture consisting in $12$ layers (as it was originally proposed by~\cite{yenduri2023generativepretrainedtransformercomprehensive}), each of which comprises $16$ attention heads, and a feed-forward layer of dimension $4096$ with Gaussian Error Linear Unit (GELU)~\cite{Hendrycks2016GaussianEL} activation function. Each attention layer is followed by a residual connection and a normalization layer to enhance training stability. This is also applied after the feed-forward layer. In order to prevent overfitting, we apply dropout with rate $0.1$ before each residual connection and normalization operations. 

The vocabulary is customized on the STL syntax, thus is limited to $35$ tokens corresponding to the numbers, the (temporal and propositional) operators, the parentheses and the blank space separating the different logical structures. Additionally, the \texttt{unk} (i.e. unknown token), \texttt{pad} (i.e. pad token), \texttt{bos} (i.e. beginning-of-sentence) and \texttt{eos} (i.e. end-of-sentence) special tokens are added to the vocabulary.

The semantic information contained in the embedding of a STL formula is integrated through the cross-attention mechanism into the generating process: during the inference phase, each auto-regressively generated token is conditioned on the information contained in this semantic embedding. 

\subsubsection{Training} 

Our experiments aim at assessing the ability of a decoder-only Transformer architecture to reconstruct a plausible formula starting from a semantic embedding. In this direction, we train from scratch the formerly described architecture using different training sets, as detailed in Section~\ref{subsec:data} and obtained different models, which we will refer to using the same name of the used training set (namely \texttt{random, hardskewed, easyskewed, balanced}).  

Another point of interest in our analysis consists in the impact of the richness of the semantic representations, i.e. the chosen embedding dimension. Indeed, we both train models with hidden dimensions of $512$ and $1024$ and studied the differences in the generalization abilities of the resulting models. This is achieved by embedding both training and test sets using the STL kernel of Equation~\ref{eq:stl-kernel}, fixing a set of either $1024$ or $512$ anchor formulae sampled from $\mathcal{F}$. As expected, the training time doubles when the hidden dimension is duplicated. To refer to the model trained with embedding of size $512$, we append the suffix \texttt{small} to the name of the model.

We train all the models for $10$ epochs (corresponding to $\sim 24000$ steps) on those training sets, all of size $78000$, with a batch size of $32$. We further elongate the training stage for the best models (according to the criteria described in Section~\ref{subsec:inversion}) for $10$ more epochs, in order to test whether or not we could observe a more refined behavior, when allowing for a greater training time. 
The optimization is performed using the AdamW optimizer~\cite{adamw}, which decouples weight decay from gradient updates for improved generalization. A linear learning rate scheduler is applied, starting with a warm-up phase of $5000$ steps before gradually decreasing over $50000$ total training steps. The initial learning rate is set to $5 \cdot 10^{-5}$, with weight decay of $0.01$ to mitigate possible overfitting.

\paragraph{At inference time} the semantic embedding of a formula is fed to the trained model along with the \texttt{bos} (begin of sentence) starting token. The model auto-regressively infers the next formula elements starting from this NeSy representation.


\section{Experiments}\label{sec:experimental}

We claim and experimentally prove the effectiveness of our methodology in capturing the semantic information contained in the STL kernel representations in two different settings: (i) approximately inverting embeddings of STL formulae, see Section~\ref{subsec:inversion} and (ii) performing the requirement mining task in several benchmark datasets, as shown in Section~\ref{subsec:reqmining}. 

\subsection{Approximately Inverting STL Kernel Embeddings}\label{subsec:inversion}

The goal of this suite of experiments is twofold: verifying that a Transformer-based model is able to effectively grasp the semantics of STL, as encoded by the STL kernel embeddings, and compare it to a related approach based on Information Retrieval (IR) techniques; investigating if and how much the model size and the training set distribution influence such capabilities. 

We test these abilities on a set of formulae with balanced depth levels, in order to robustly assess the aforementioned \textit{desiderata}. The depths of the formulae range from $2$ to $7$, and it is worth noting that depth-$4$ formulae can already involve up to three different temporal operators, making them quite complex. 

As an example, we can consider the sentence ``the temperature $\tau$ of the room will reach $25$ degrees within the next $10$ minutes and will stay above $22$ degrees for the successive $60$ minutes", which translates in STL as  $F_{[0, 10]} (\tau \geq 25 \wedge G_{[0, 60]} \tau \geq 22)$ (i.e. a requirement with depth $4$ and $3$ logical operators). 

Ideally, if we query our model with the embedding $k(\varphi)$ of a known formula $\varphi$, then we expect to decode a specification $\hat{\varphi}$ which has the same semantics of $\varphi$, with possibly a different syntax. To verify this property, we can consider the robustness vector of a formula, i.e. $\bm{\rho}(\varphi) = [\rho(\varphi, \xi_i)]_{i=1}^M$ for an arbitrary, but fixed, set of trajectories $\Xi = \{\xi_i\}_{i=1}^M$, and compute the quantity $d(\bm{\rho}(\varphi), \bm{\rho}(\hat{\varphi})) = ||\bm{\rho}(\varphi) - \bm{\rho}(\hat{\varphi})||_2$: given that two STL formulae are semantically similar if they behave similarly on the same set of signals, a lower value of $d(\bm{\rho}(\varphi), \bm{\rho}(\hat{\varphi}))$ indicates a good approximation of the inverse of $k(\varphi)$. Following the same reasoning line, as additional metrics we can consider: (a) the cosine similarity between robustness vectors, namely $\text{cos}(\bm{\rho}(\varphi), \bm{\rho}(\hat{\varphi})) = \frac{\bm{\rho}(\varphi) \cdot \bm{\rho}(\hat{\varphi})}{||\bm{\rho}(\varphi)|| ||\bm{\rho}(\hat{\varphi})||}\in [-1, 1]$, with $\text{cos}(\bm{\rho}(\varphi), \bm{\rho}(\hat{\varphi})) = 1$ if the original $\varphi$ and reconstructed $\hat{\varphi}$ are (un)satisfied on the same set of trajectories of $\Xi$, possibly with different robustness degrees and (b) average number of signals in which $\varphi$ and $\hat{\varphi}$ have opposite satisfaction status, i.e. $\text{diff}(\bm{s}(\varphi), \bm{s}(\hat{\varphi})) = \frac{\sum_{i=1}^M \mathbb{I}(s(\varphi, \xi_i) \neq s(\hat{\varphi}, \xi_i))}{M}$ being $\mathbb{I}$ the indicator function. When $\text{diff}(\bm{s}(\varphi), \bm{s}(\hat{\varphi})) = 0$ the the original $\varphi$ and reconstructed $\hat{\varphi}$ are (un)satisfied on exactly the same set of trajectories. 

Besides comparing all the trained models on the above mentioned metrics, we also analyze their performance w.r.t. the IR-based approach devised in~\cite{semantic-db}, in which a dense vector database (hereafter denoted as DB) containing STL kernel embeddings of millions of formulae is built, so that an approximate inverse $\hat{\varphi}$ of the embedding $k(\varphi)$ of a specification $\varphi$ is obtained with approximate nearest neighbors by querying the DB with $k(\varphi)$.

\begin{figure}[t]
    \centering
    \begin{minipage}{0.44\textwidth}
        \includegraphics[width=1\linewidth]{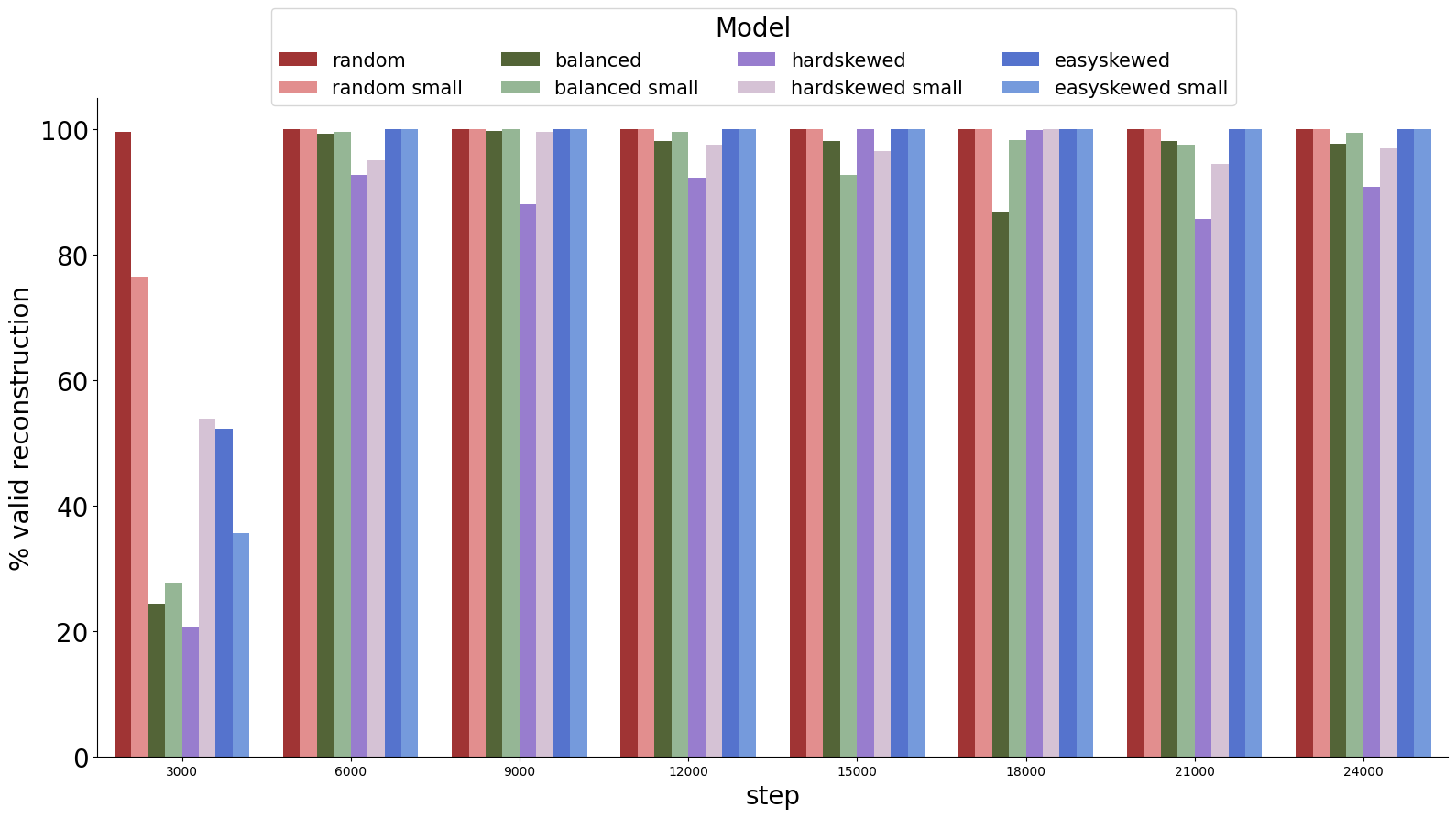}
    \end{minipage}
    \hfill
    \begin{minipage}{0.54\textwidth}
        \centering
        \includegraphics[width=\linewidth]{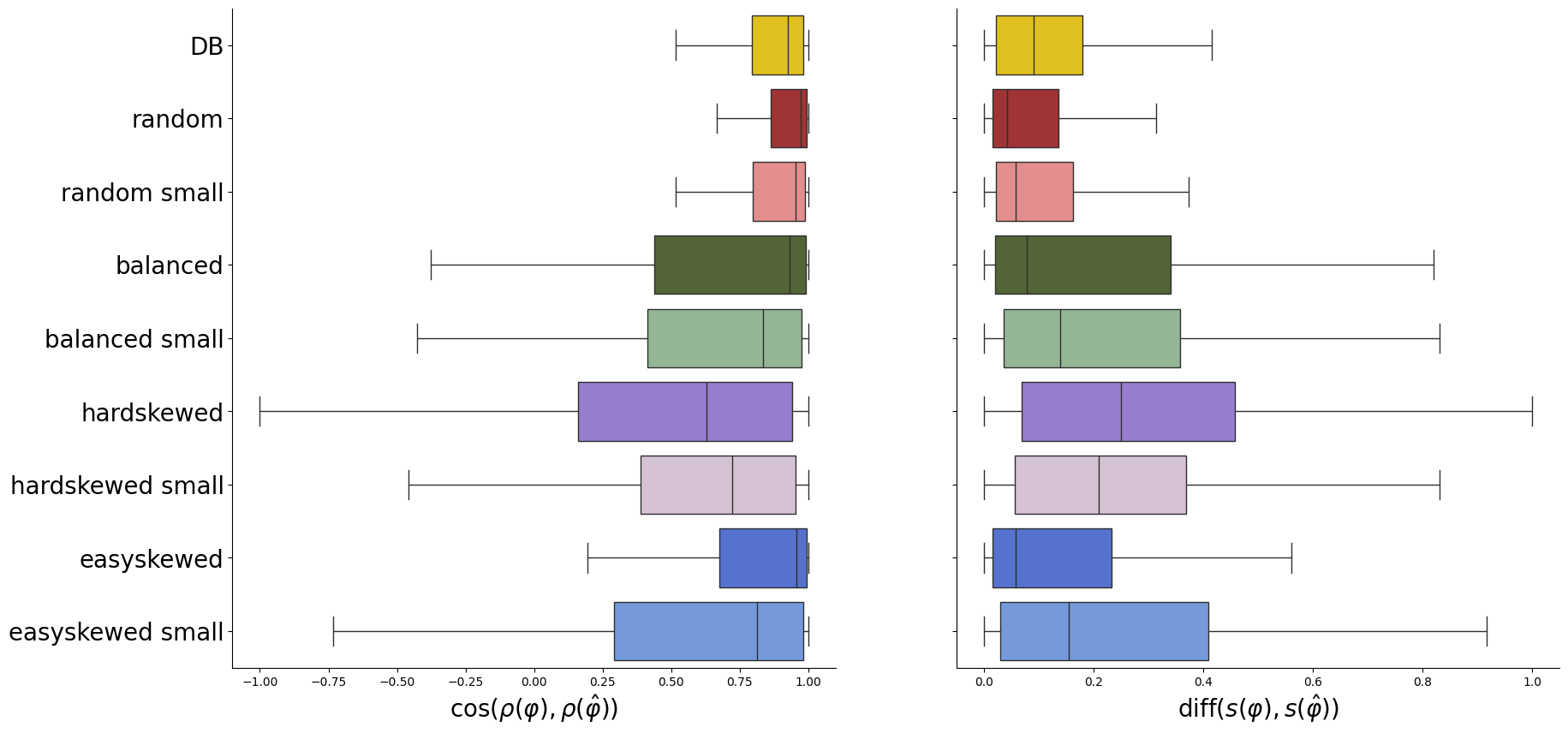}
    \end{minipage}
    \caption{Considering all the trained models: (left) percentage of syntactically valid formulae generated across the training steps and (right) comparison of results between transformer-based after $24000$ steps of training models and semantic vector database.}
    \label{fig:validity-comparison}
\end{figure}

\begin{table}[t]
\caption{Comparison of results of different checkpoints of the best Transformer-based models and of the DB. On parenthesis the percentile of the distribution of the corresponding metric on a random set of formulae.}
\label{tab:best-results-percentile}
\centering
\resizebox{1.\linewidth}{!}{
\begin{tabular}{l|llll|llll|llll}
\toprule
{} & \multicolumn{4}{c}{$d(\bm{\rho}(\varphi), \bm{\rho}(\hat{\varphi}))$} & \multicolumn{4}{c}{$\text{cos}(\bm{\rho}(\varphi), \bm{\rho}(\hat{\varphi}))$} & \multicolumn{4}{c}{$\text{diff}(\bm{s}(\varphi), \bm{s}(\hat{\varphi}))$}\\
\midrule
Model &  $1$quart  &   median &   $3$quart &  $99$perc &  $1$quart &   median &   $3$quart &  $99$perc & $1$quart &   median &   $3$quart &  $99$perc \\
 DB & $16.51$ $(1)$ & $28.13$ $(4)$ & $43.87$ $(12)$ & $81.46$ $(75)$ & $0.795$ $(91)$ & $0.9262$ $(98)$ & $0.9802$ $(99)$ & $0.9995$ $(100)$ & $0.0225$ $(1)$ & $0.0908$ $(4)$ & $0.1797$ $(11)$ & $0.4831$ $(47)$ \\
\texttt{random} (step $24$K) & $8.447$ $(1)$ & $17.63$ $(2)$ & $36.03$ $(7)$ & $123.4$ $(100)$ & $0.8617$ $(95)$ & $0.9718$ $(99)$ & $0.9939$ $(100)$ & $1.000$ $(100)$ & $0.0159$ $(1)$ & $0.0423$ $(2)$ & $0.1354$ $(7)$& $0.7571$ $(85)$ \\
\texttt{random small} (step $24$K) & $11.66$ $(1)$ & $23.06$ $(2)$ & $44.51$ $(12)$ & $113.4$ $(100)$ & $0.7985$ $(91)$ & $0.9518$ $(99)$ & $0.9885$ $(100)$ & $1.000$ $(100)$ & $0.0224$ $(1)$ & $0.0586$ $(3)$ & $0.1631$ $(9)$ & $0.7154$ $(81)$\\
\texttt{random} (step $48$K)  & $6.054$ $(0)$ & $12.71$ $(1)$ & $29.89$ $(4)$ & $110.1$ $(100)$ & $0.9012$ $(96)$ & $0.9854$ $(100)$ & $0.9971$ $(100)$ & $1.000$ $(100)$ & $0.0073$ $(0)$ & $0.0274$ $(1)$ & $0.1085$ $(6)$ & $0.6625$ $(75)$\\
\texttt{random small} (step $48$K)  & $7.462$ $(1)$ & $16.01$ $(1)$ & $31.22$ $(5)$ & $101.1$ $(100)$ & $0.8944$ $(96)$ & $0.9774$ $(99)$ & $0.9952$ $(100)$ & $1.000$ $(100)$ & $0.0106$ $(1)$ & $0.0359$ $(2)$ & $0.1159$ $(6)$ & $0.5563$ $(61)$ \\
\bottomrule
\end{tabular}
}
\end{table}

In Figure~\ref{fig:validity-comparison} and Table~\ref{tab:best-results-percentile} we report the results of tests done on a \texttt{balanced} test set of $3000$ formulae, i.e. constructed as the one used to train the homonymous model described in Section~\ref{subsec:data}. From Figure~\ref{fig:validity-comparison} (left) it is possible to notice that \texttt{random} model decodes only syntactically valid formulae after just $3000$ steps ($\sim 1.5$ epoch) of training, and that by step $24000$ ($\sim 10$ epochs) all architectures reach $\geq 85\%$ of valid generations, highlighting the ability of Transformer-based models to grasp the syntax of STL. Moreover, on Figure~\ref{fig:validity-comparison} (right) we can check that the two models trained on the \texttt{random} training set are outperforming all the others, with the bigger one surpassing also the DB approach on the tested metrics. Indeed, after $24000$ steps they achieve a median $\text{cos}(\bm{\rho}(\varphi), \bm{\rho}(\hat{\varphi}))\geq 0.95$ and median $\text{diff}(\bm{s}(\varphi), \bm{s}(\hat{\varphi}))\leq 0.06$. Pushing further the investigation, we trained these $2$ top-performing model for additional $24000$ steps (hence for a total of $\sim 20$ epochs) and showed the results on Table~\ref{tab:best-results-percentile}. There, we not only report the quantiles of the test distribution of the $3$ considered metrics, but also the corresponding percentile of the distribution of the same quantities computed on a set of $10000$ random formulae. Hence, we see that the additional training steps bring a relatively small improvement on all the considered metrics, and that the Transformer-based architecture is able to grasp the semantics of STL after $\sim 10$ epochs of training,  reporting a median $\sim0.98$ cosine similarity between the vectors of the robustness of the ground truth and retrieved formula, with the two having a different satisfaction status on $\sim3\%$ of trajectories. The median value of $d(\bm{\rho}(\varphi), \bm{\rho}(\hat{\varphi}))$, despite not being interpretable, corresponds to the $1^{\text{st}}$ percentile of the random distribution, hence significantly low, enforcing the observation that the $\texttt{random}$ model effectively learns the semantics of STL as encoded by the STL kernel embeddings. 

\begin{figure}[t]
    \centering
    \begin{minipage}{0.44\textwidth}
        \includegraphics[width=1\linewidth]{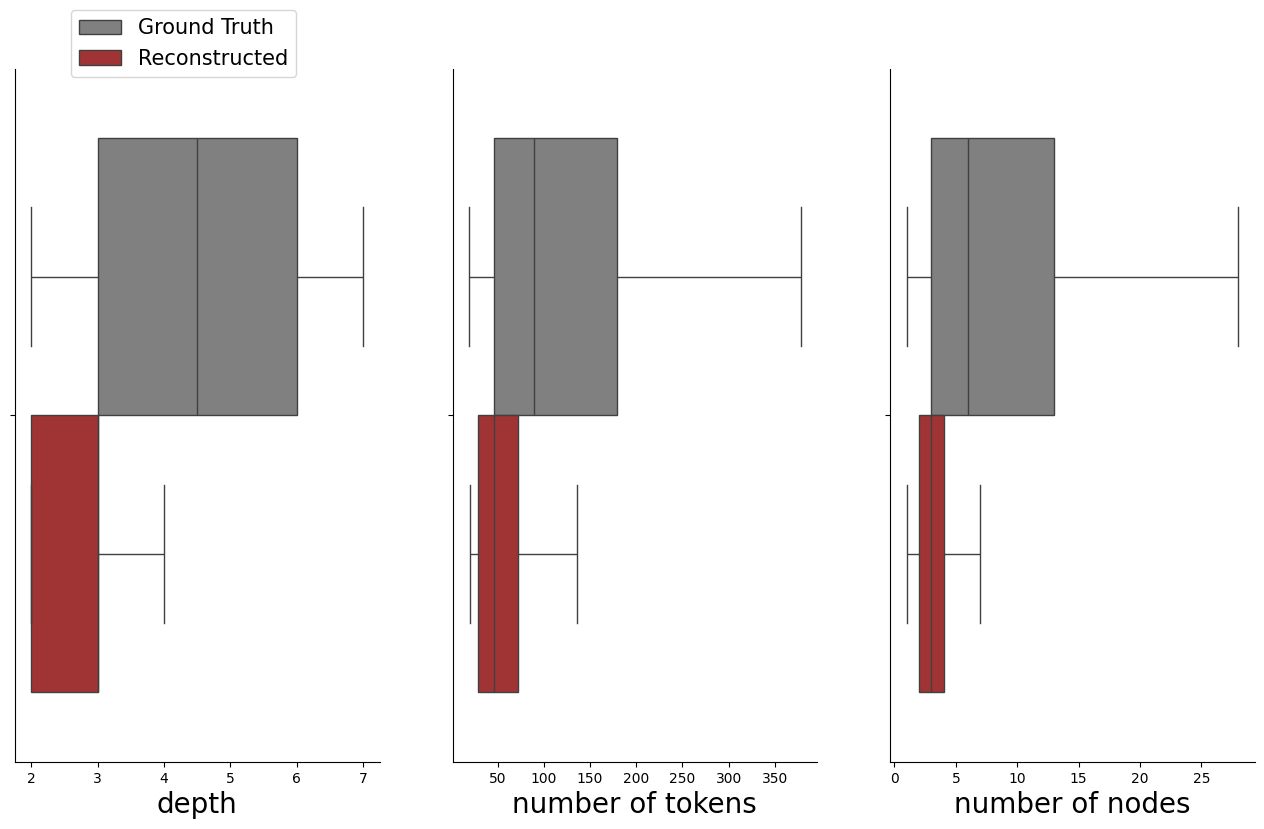}
    \end{minipage}
    \hfill
    \begin{minipage}{0.55\textwidth}
        \centering
        \includegraphics[width=\linewidth]{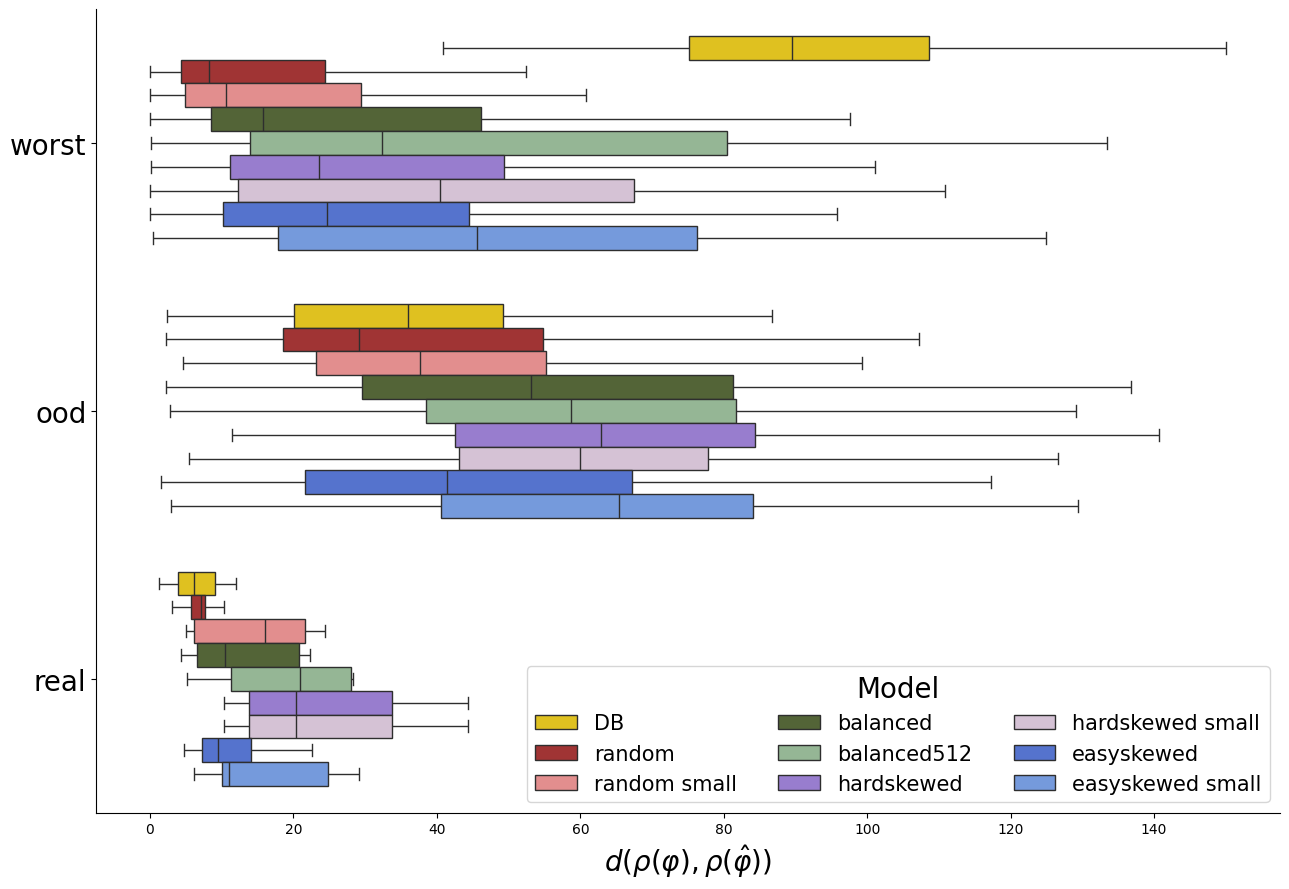}
    \end{minipage}
    \caption{(left) Distribution of depth, number of tokens and number of nodes of the ground truth and reconstructed formulae (by the \texttt{random} architecture) composing the \texttt{balanced} test set; (right) comparison of results between
transformer-based models and semantic vector database on the test sets \texttt{worst, ood, real}.}
    \label{fig:structure-ood}
\end{figure}


Notably, from Figure~\ref{fig:structure-ood}, we observe that the \texttt{random} model tends to decode formulae that are syntactically simpler than those used to generate the test embeddings. We argue that this behavior may stem from the high syntactic variability observed in its training set, as shown in Figure~\ref{fig:structure-training}. Since formulae with very different syntactic structures might nonetheless share very similar semantics, it is likely that the \texttt{random} training set contains pairs of specifications that are semantically close but very different in their structure, in terms e.g. of depth and number of nodes. However, given that the majority of training requirements are relatively simple, the Transformer is biased toward decoding shorter formulae. This results in a significant advantage for downstream applications in terms of interpretability: we are able to reconstruct a formula with very similar semantic meaning yet syntactically simpler.

Additionally, we tested our model on the following datasets: 
\begin{itemize}
    \item \texttt{worst}: a set of $400$ formulae sampled from $\mathcal{F}$ using $p_{\text{leaf}} = 0.4$ which we experimentally found to be those on which the DB performs the worst. This set aims at testing whether the transformer-based model is able to generalize to the space of STL kernel embeddings which are scarcely covered by the formulae contained in DB; 
    \item \texttt{ood}: a set of $500$ STL formulae of depth $8$, which are absent from the training sets of the \texttt{balanced}, \texttt{easyskewed}, and \texttt{hardskewed} models, and only present in a very limited amount in that of the \texttt{random} models; 
    \item \texttt{real}: a small set of $15$ formulae taken from~\cite{archcomp}, to check whether the proposed model is effective on requirements coming form real CPS applications. 
\end{itemize}

The results are shown in Figure~\ref{fig:structure-ood} (right), where all models successfully decode only valid formulae. The x-axis represents $d(\bm{\rho}(\varphi), \bm{\rho}(\hat{\varphi}))$, the distance between the robustness vectors of the gold and reconstructed formulae, and should be interpreted as \emph{the lower, the better}. While this metric may offer limited interpretability, it allows for a more fine-grained assessment of the relative differences between formulae.

For what concerns the \texttt{worst} dataset, all the trained model outperform the DB, confirming that the transformer-based decoder is able to effectively grasp the semantic of STL. Regarding the \texttt{ood} set, we notice that the DB is the best performing architecture, with the two models trained on the \texttt{random} set showing comparable performance, and the others reporting slightly worst results. This indicates that a greater variability in the structure of formulae over which the models are trained aids in length generalization. Finally, the \texttt{random} model is best performing on the \texttt{real} dataset, immediately followed by the DB.

\subsection{STL Requirement Mining}\label{subsec:reqmining}

The objective of the experiments described in this Section is that of checking whether our model can be leveraged  for solving the requirement task, i.e. that of inferring a property (in the form of STL formula) characterizing the behavior of an observed system. We follow the approach of~\cite{enumerate-stl,roge,semantic-db,gp-one,gp-two,decision-tree} and frame it as a supervised two-class classification problem, where the input consists of a set of trajectories divided in: those classified as regular (or positive) and those identified as anomalous (or negative), denoted as $\mathcal{D}_p$ and $\mathcal{D}_n$, respectively. The output is a (set of) STL formula(e) designed to distinguish between these two subsets. Additionally, we assume that these datasets originate from unknown stochastic processes, denoted as $\bm{X}_p$ and $\bm{X}_n$, respectively and adopt the approach of \cite{roge} and maximize the following function in order to mine a STL formula $\varphi$ able to discriminate between positive and negative trajectories:
\begin{equation}
    G(\varphi) = \frac{\mathbb{E}_{\mathbf{\bm{X}_p}}[R_{\varphi}(\mathbf{\bm{X}_p})]- \mathbb{E}_{\mathbf{\bm{X}_n} }[R_{\varphi}(\mathbf{\bm{X}_n})]}{\sigma_{\mathbf{\bm{X}_p}}(R_{\varphi}(\mathbf{\bm{X}_p})) + \sigma_{\mathbf{\bm{X}_n}}(R_{\varphi}(\mathbf{\bm{X}_n}))}
    \label{eq:reqmining-optim}
\end{equation}
denoting as $\mathbb{E}_{\bm{X}}[R_{\varphi}(\bm{X})]$ and $\sigma_{\bm{X}}[R_{\varphi}(\bm{X})]$ respectively the expected value and the standard deviation of the robustness of a formula $\varphi$ on trajectories sampled from the system $\bm{X}$. Following~\cite{semantic-db}, we (i) frame the learning problem as the optimization of Equation~\ref{eq:reqmining-optim} in the latent semantic space of formulae, i.e. in the space of embeddings of formulae individuated by the STL kernel of Equation~\ref{eq:stl-kernel} and (ii) tackle it by means of Bayesian Optimization (BO), and more specifically of the Gaussian Process Upper-Confidence Bound (GP-UCB) algorithm~\cite{gp-ucb}.
Hence, the overall iterative methodology follows these steps in each iteration:
\begin{enumerate}
    \item Optimize the acquisition function based on the current set of pairs $(k(\varphi), G(\varphi))$ of kernel embeddings and corresponding objective function values to obtain new candidate embeddings;
    \item Retrieve STL formulae corresponding to such candidate embeddings by inverting the them using the transformer-based decoder model;
    \item Evaluate the objective function $G(\varphi)$ on the obtained formulae and record the pair achieving the maximum value.
\end{enumerate}

\begin{table}[t]
\caption{Best mined formula $\hat{\varphi}$, mean and standard deviation of $MCR$ and $Prec$ in the test set across $3$ different initialization seeds.}
\label{tab:mining-results}
\centering
\resizebox{\linewidth}{!}{
\begin{tabular}{ll|lll}
\toprule
{} & {} & $\hat{\varphi}$ & $MCR$ & $Prec$ \\
\midrule
\multirow{4}{*}{\texttt{Linear}} 
&\texttt{random} (step $24$K) & $G_{[3, \infty]} (x_0\geq -11.44)$ & $0.0200 \pm 0.0291$ & $0.9650 \pm 0.0489$ \\
& \texttt{random small} (step $24$K) & $G(x_0\geq -15.09)$ & $0.0150 \pm 0.0300$ & $0.9714 \pm 0.0571$  \\
& \texttt{random} (step $48$K) & $F (x_0\geq 0.1918)$ & $0.0150 \pm 0.0300$ & $0.9700 \pm 0.0600$ \\
& \texttt{random small} (step $48$K) & $G (x_0\geq -5.586)$ & $0.0150 \pm 0.0201$ & $0.9670 \pm 0.0485$ \\
\midrule
\multirow{4}{*}{\texttt{HAR}} 
& \texttt{random} (step $24$K) & $G_{[5, \infty]} (x_0\geq 39.93)$ & $0.1333 \pm 0.2666$ & $0.8666 \pm 0.2668$  \\
& \texttt{random small} (step $24$K) & $G_{[3, 8]} (x_0\geq 29.37)$ & $0.1368 \pm 0.2736$ & $0.8631 \pm 0.2736$  \\
& \texttt{random} (step $48$K) & $G_{[2, \infty]} (x_0\geq 33.62)$ & $0.0947 \pm 0.1894$ & $0.9052 \pm 0.1894$  \\
& \texttt{random small} (step $48$K) & $G_{[5, 9]} (x_0\geq 32.92)$ & $0.0736\pm 0.1473$ & $0.9304 \pm 0.1392$  \\
\midrule
\multirow{4}{*}{\texttt{LP$5$}} 
& \texttt{random} (step $24$K) & $G (x_2\geq -9.272)$ & $0.0571 \pm 0.0534$ & $0.8818 \pm 0.0754$  \\
& \texttt{random small} (step $24$K) & $G (x_2\geq -12.75)$ & $0.1713 \pm 0.2731$ & $0.9377 \pm 0.0509$  \\
& \texttt{random} (step $48$K) & $G (x_2\geq -8.142)$ & $0.0857 \pm 0.0534$ & $0.9043 \pm 0.0830$  \\
& \texttt{random small} (step $48$K) & $G (x_2\geq -5.614)$ & $0.0857 \pm 0.0699$ & $0.8777 \pm 0.0976$  \\
\midrule
\multirow{4}{*}{\texttt{Train}} 
& \texttt{random} (step $24$K) & $F_{[18, 22]} (G_{[13, 16]} (x_0\leq 4.823))$ & $0.0468 \pm 0.0422$ & $0.9183 \pm 0.0716$  \\
& \texttt{random small} (step $24$K) & $G_{[16, 19]} (F_{[12, 18]} (x_0\leq 1.894) )$ & $0.0428 \pm 0.0346$ & $0.7979 \pm 0.3113$\\
& \texttt{random} (step $48$K) & $G_{[13, \infty]} (x_0\leq 24.64)$ & $0.0652 \pm 0.0717$ & $0.8968 \pm 0.1046$ \\
& \texttt{random small} (step $48$K) & $\lnot (F_{[16, \infty]} (x_0\geq 1.538))$ & $0.1101 \pm 0.1593$ & $0.7929 \pm 0.3041$  \\
\bottomrule
\end{tabular}
}
\end{table}

In our experiments, we employ Gaussian Processes with a Matérn kernel, initialized with a batch of $100$ points. Due to the high-dimensional STL kernel space (either $1024$ or $512$, as detailed in Section~\ref{subsec:architecture}), we optimize the Upper Confidence Bound (UCB) acquisition function via Stochastic Gradient Descent (SGD). 

Following the related literature~\cite{enumerate-stl,semantic-db}, we test the Linear System (\texttt{Linear}), Human Activity Recognition (\texttt{HAR}), Robot Execution Failure in Motion with Part (LP$5$) and Cruise Control of Train (\texttt{Train}) time series classification datasets. Results in terms of Misclassification Rate ($MCR$) and Precision ($Prec$), are reported in Table~\ref{tab:mining-results} for the most promising checkpoints of our model; recall is not shown being always equal to $1$. For all datasets the \texttt{HAR} the \texttt{random} model trained for $24000$ steps achieves an accuracy $\geq 90\%$ (which is always achieved by the \texttt{random} model trained for $48000$ steps). Additionally, in Table~\ref{tab:mining-results} we show the best mined formula $\hat{\varphi}$ by each architecture, which we choose as the one obtaining the lowest MCR (namely $0$ in all reported cases), using the syntactic simplicity as tiebreaker. Notably, in line with recent related works~\cite{enumerate-stl,semantic-db}, all $\hat{\varphi}$ have at most $3$ nodes in their syntax tree, hence are highly interpretable, thus promoting knowledge discovery of the resulting system. 

Since our architecture is a learned model, it is highly likely that, when queried with a vector that is out-of-distribution (OOD) with respect to the STL kernel embeddings, it produces an invalid formula. This situation can arise, for example, during the exploration phase of the UCB acquisition function. In such cases, we assign an extremely low value to the objective function in Equation~\ref{eq:reqmining-optim}, in order to encourage the GP-UCB to explore more plausible regions of the STL embedding space.
On the one hand, this has the positive effect of automatically detecting OOD samples. A possible interpretation of why this occurs is that the embeddings of logical formulae lives in a lower dimensional manifold, hence vectors outside this manifold do not correspond to real requirements, hence embeddings corresponding to such a situation (or belonging to formulae which are semantically very different from the training data) are likely to give rise to invalid outputs. 
More in detail, the $1024$-dimensional architectures take $\sim1$s to invert an embedding, while the $512$-dimensional roughly $0.5$s. Practically, due to the mentioned issue, we witness an computational time of $\sim 300$s for the bigger models and $\sim 120$s for the smaller ones, on the tests reported in Table~\ref{tab:mining-results}.

\section{Conclusion}

In this work, we investigate the possibility of using a decoder-only Transformer-based architecture to invert continuous representation of Signal Temporal Logic formulae into valid requirements, which are semantically similar to the ground truth searched formulae. Our experiments prove that such model is able to generate valid formulae after only $1$ epoch of training and to generalize to the semantics of the logic in about $10$ epochs. Indeed, not only it is able to decode a given embedding into formulae which are semantically close to gold references, but that are often simpler in terms of length and nesting.
It is worth noting that, due to the nature of Transformer-based architectures (and neural models in general), deriving meaningful theoretical bounds with practical applicability is extremely challenging. There is an inherent trade-off between speed and accuracy on one hand, and strong theoretical guarantees on the other.

These results leave open the question of whether Transformer-based models can be leveraged in other related Neuro-Symbolic tasks, possibly involving different formal languages, such as first-order logic. We indeed envision that such powerful generative capabilities of Language Models might be leveraged both, as we do here, to decode formulae from a continuous vector representing their semantic and, in an encoder-decoder setting, even to devise invertible by-design continuous semantic-preserving representations, opening the doors to a whole new range of applications.  
Finally, a further direction that could be explored involves the interpretability of these models; in particular, it could be interesting to study their internals, and check e.g. how the attention heads contribute in the decoding process when provided with the semantic embedding. 


\begin{credits}
\subsubsection{\ackname} 

This study was carried out within the PNRR research activities of the consortium iNEST (Interconnected North-Est Innovation Ecosystem) funded by the European Union Next-GenerationEU (Piano Nazionale di Ripresa e Resilienza (PNRR) – Missione 4 Componente 2, Investimento 1.5 – D.D. 1058 $\cdot$ 23/06/2022, ECS\_00000043). 
This manuscript reflects only the Authors’ views and opinions, neither the European Union nor the European Commission can be considered responsible for them.
Gabriele Sarti is supported by the Dutch Research Council (NWO) for the project InDeep (NWA.1292.19.399).

\subsubsection{\discintname}
The authors have no competing interests to declare that are
relevant to the content of this article. 
\end{credits}
%
%
%
%

\end{document}